%% file: eccv2018submission.tex
\documentclass[runningheads]{llncs}
\usepackage{graphicx}
\usepackage{amsmath,amssymb} 
\usepackage[usenames, dvipsnames]{color}
\usepackage[width=122mm,left=12mm,paperwidth=146mm,height=193mm,top=12mm,paperheight=217mm]{geometry}
\usepackage{caption}
\usepackage{subcaption}
\usepackage{amsmath}
\usepackage{amssymb}
\usepackage{mathtools}
\usepackage{xspace}
\makeatletter
\DeclareRobustCommand\onedot{\futurelet\@let@token\@onedot}
\def\@onedot{\ifx\@let@token.\else.\null\fi\xspace}

\def\eg{\emph{e.g}\onedot} 
\def\ie{\emph{i.e}\onedot}

\def\etal{\emph{et al}\onedot}

\makeatother
\begin{document}
\pagestyle{headings}
\mainmatter
\def\ECCV18SubNumber{6}  

\title{Action Anticipation By Predicting Future Dynamic Images} 

\titlerunning{Action Anticipation By Predicting Future Dynamic Images}

\authorrunning{C. Rodriguez et al.}

\author{Cristian Rodriguez, Basura Fernando and Hongdong Li}
\institute{Australian Centre for Robotic Vision, ANU, Canberra, Australia\\ \texttt{\{cristian.rodriguez, basura.fernando, hongdong.li\}@.anu.edu.au} }

\maketitle

\begin{abstract}
Human action-anticipation methods predict what is the future action by observing only a few portion of an action in progress.
This is critical for applications where computers have to react to human actions as early as possible such as autonomous driving, human-robotic interaction, assistive robotics among others.
In this paper, we present a method for human action anticipation by predicting the most plausible future human motion.  
We represent human motion using {\em Dynamic Images} \cite{bilen2016dynamic} and make use of tailored loss functions to encourage a generative model to produce accurate future motion prediction. 
Our method outperforms the currently best performing action-anticipation methods by 4\% on JHMDB-21, 5.2\% on UT-Interaction and 5.1\% on UCF 101-24 benchmarks.
\end{abstract}

\section{Introduction}
\input{intro}

\section{Related work}
\input{related}
\section{Method}
\input{method}
\section{Experiments and results}
\input{experiments}
\section{Discussion}
\input{discussion}

\bibliographystyle{splncs}
\bibliography{egbib}
\end{document}

%% file: intro.tex
When interacting with other people, human beings have the ability to anticipate the behaviour of others and act accordingly. This ability comes naturally to us and we make use of it subconsciously. Almost all human interactions rely on this {\em action-anticipation} capability. For example, when we greet each other, we tend to anticipate what is the most likely response and act slightly proactively. When driving a car, an experienced driver can often predict the behaviour of other road users. Tennis players predict the trajectory of the ball by observing the movements of the opponent. The ability to anticipate the action of others is essential for our social life and even survival. It is critical to transfer this ability to computers so that we can build smarter robots in the future, with better social interaction abilities that think and act fast.

In computer vision, this topic is referred to as {\em action anticipation}~\cite{ma2016learning,ryoo2011human,aliakbarian2017encouraging,soomro2016online,soomro2016predicting} or early action prediction~\cite{lan2014hierarchical,yu2012predicting}. 
Although action anticipation is somewhat similar to {\em action recognition}, they differ by the information being exploited.  Action-recognition processes the entire action within a video and generate a category label, whereas action-anticipation aims to recognise the action {\em as early as possible}.  More precisely, action-anticipation needs to predict the future action labels as early as possible by processing fewer image frames (from the incoming video), even if the human action is still in progress. 

Instead of directly predicting action labels~\cite{aliakbarian2017encouraging}, we propose a new method that generates future motion representation from partial observations of human action in a video. We argue that the generation of future motion representation is more intuitive task than generating future appearance, hence easier to achieve. A method that is generating future appearance given the current appearance requires to learn a conditional distribution of factors such as colour, illumination, objects and object parts, therefore, harder to achieve.  In contrast, a method that learns to predict future motion does not need to learn those factors.  Furthermore, motion information is useful for recognising human actions~\cite{Bilen2017,Simonyan2014} and can be presented in various image forms~\cite{Bilen2017,Ahad2012}.

In this paper we propose to predict future motion representation for action anticipation. 
Our method hallucinates what is in the next motion representation of a video sequence given only a fraction of a video depicting a partial human action. 
We make use of a convolutional autoencoder network that receives a motion image as input at time $t$ and outputs a motion image for the future (\eg $t+1$). 
Using Markov assumption, we generate more motion images of the future using already generated motion images (\ie we generate motion images for time $t+1, \cdots, t+k$).
Then we process generated motion images using Convolutional Neural Network (CNN) to make action predictions for the future.
As we are able to generate future motion images, now we are able to predict human actions only observing few frames of a video containing an action.  

We train our action anticipation and motion generation network with several loss functions.
These loss functions are specifically tailored to generate accurate representations of future motion and to make accurate action predictions.
\begin{figure}[t]
    \centering
    \includegraphics[width=0.49\textwidth]{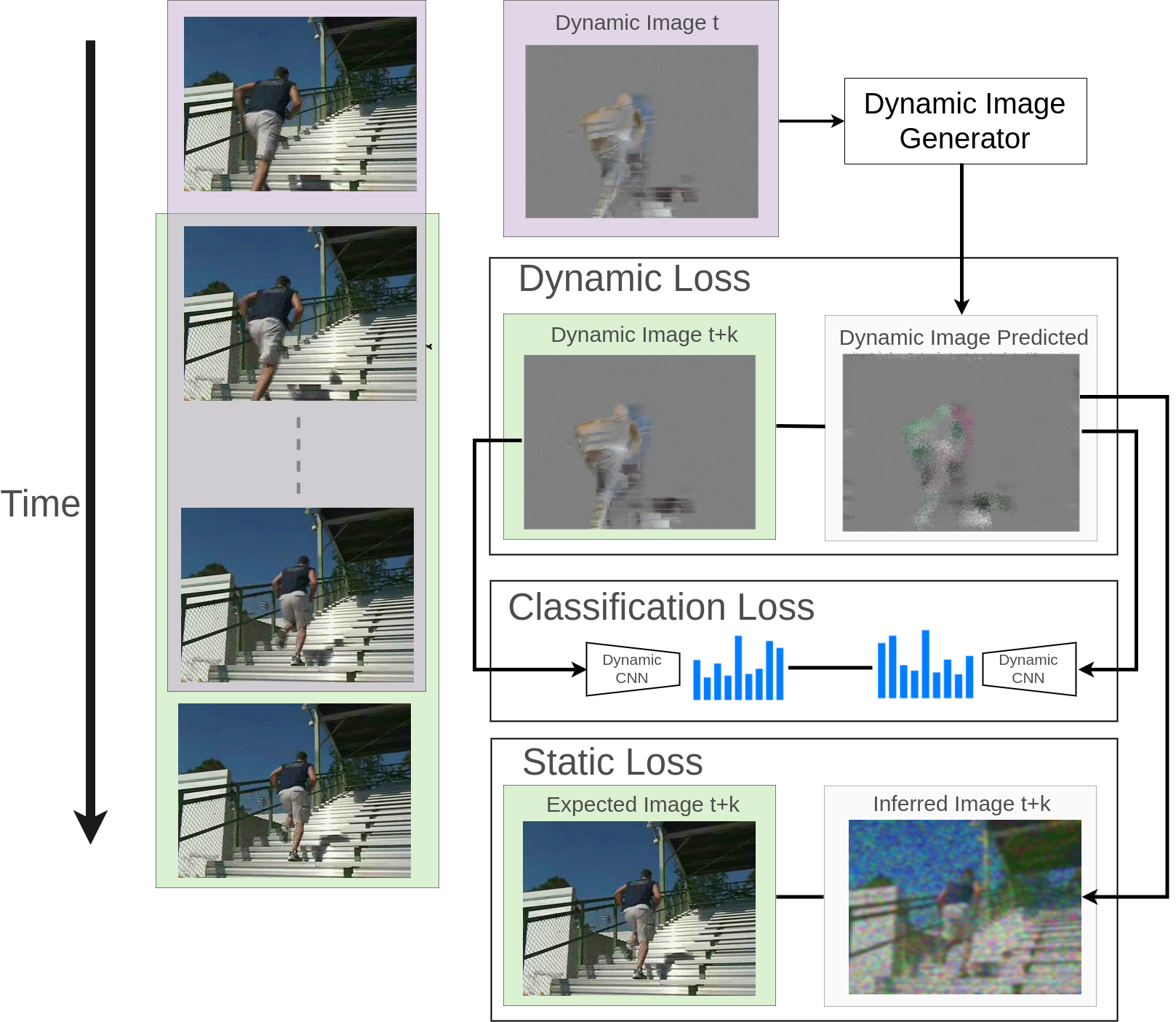}
    \caption{Training of our generation module using multiple loss functions. \textbf{a) \textit{Dynamic Loss}} evaluates the difference in motion information between predicted and ground truth dynamic image using $\mathcal{L}_2$ norm. \textbf{b) \textit{Classification Loss}} takes care of generating dynamic images that are useful for action anticipation. \textbf{c) Static Loss} computes the $\mathcal{L}_2$ norm between predicted and ground truth RGB information at $t+k$ to evaluate the difference in appearance.} 
    \label{fig:losses}
\end{figure}

Clearly, the motion information depends on the appearance and vice versa. 
For example, motion representations such as the optical flow relies on two consecutive RGB frames. 
Similarly, the content of dynamic images~\cite{Bilen2017} relies on the appearance of consecutive frames.
The relationship between static appearance and motion information is somewhat surprising and mysterious~\cite{Carreira2017}. 
Recently, proposed dynamic images has managed to explore this relationship to some degree of success~\cite{Bilen2017}. 
In particular, dynamic images summarise the temporal evolution of appearance of few frames (\eg 10 frames) into a single image.
Therefore, this motion summary image (a.k.a. dynamic image) captures the motion information of those frames.
In this work, we hallucinate dynamic images for the future and use them for the task of action anticipation
\footnote{However, the main concept of this paper is applicable for other types of motion images as well (optical flow, motion history images).}.

We generate dynamic images using both expected appearance and motion of the future.
Specifically, future dynamic images are generated by taking into account both reconstructive loss (coined \emph{dynamic loss}) and future expected appearance which is coined \emph{static loss}.
As motion and appearances should adhere to each other, static loss is designed to satisfy expected future appearance in the generated
dynamic images.
In addition to that our generated dynamic images make use of class information and therefore discriminative.
These loss functions are tailored to generate accurate future dynamic images as is depicted in Fig.~\ref{fig:losses}.
In a summary, we make the following contributions:
\begin{itemize}
 \item Using a simple CNN architecture, we demonstrate the effectiveness of dynamic images for future content prediction.
 \item We design a set of effective loss functions to produce accurate future dynamic images.
 \item We obtain state-of-the-art performance for early activity recognition on standard benchmarks.
\end{itemize}

%% file: related.tex
Action prediction and anticipation literature can be classified into deep learning and non-deep learning-based methods. 

Human activity prediction is studied using integral histograms of spatial-temporal bag-of-features coined dynamic bag-of-words in the early days~\cite{ryoo2011human}. Yu~\etal~\cite{Yu2012} propose to use spatial-temporal action matching for early action prediction task using spatial-temporal implicit shape models. Li~\etal~\cite{Li2014}, propose to explore sequence mining where a series of actions and object co-occurrences are encoded as symbolic sequences. Kong~\etal~\cite{kong2014discriminative} explore the temporal evolution of human actions to predict the class label as early as possible. This model~\cite{kong2014discriminative} captures the temporal dynamics of human actions by explicitly considering all the history of observed features as well as features in smaller temporal segments. More recently, Soomro~\etal~\cite{soomro2016predicting} propose to use binary SVMs to localise and classify video snippets into sub-action categories and obtain the final class label in an online manner using dynamic programming. Because it is needed to train one classifier per sub-action,~\cite{soomro2016online} extended this approach using a structural SVM formulation. Furthermore, this method introduces a new objective function to encourage the score of the correct action to increase as time progresses~\cite{soomro2016online}.

While all above methods utilise handcrafted features, most recent methods use deep learning approaches for action anticipation~\cite{ma2016learning,aliakbarian2017encouraging,vondrick2016anticipating}. Deep learning-based methods can be primarily categorised into two types; 1. methods that rely on novel loss functions for action anticipation~\cite{ma2016learning,aliakbarian2017encouraging,jain2016recurrent} and 2. methods that try to generate future content by content prediction~\cite{vondrick2016anticipating}.

In this context,~\cite{ma2016learning} propose to use a Long Short-Term Memory (LSTM) with ranking loss to model the activity progression and use that for effective action prediction task. They use Convolutional Neural Network (CNN) features along with a LSTM to model both spatial and temporal information. Similarly, in~\cite{jain2016recurrent}, a new loss function known as the exponentially growing loss is proposed. It tries to penalize errors increasingly over time using a LSTM-based framework. Similarly, in~\cite{aliakbarian2017encouraging}, a novel loss function for action anticipation that aims to encourage correct predictions as early as possible is proposed. The method in~\cite{aliakbarian2017encouraging} tries to overcome ambiguities in early stages of actions by preventing false negatives from the beginning of the sequence. Furthermore, a recently online action localisation method is presented which can also be used for online early action predictions~\cite{Singh2017}. However, this method primarily focuses on online action detection.

Instead of predicting the future class label, in~\cite{vondrick2016anticipating}, the authors propose to predict the future visual representation. However, the main motivation in~\cite{vondrick2016anticipating} is to learn representations using unlabeled videos. Our work is different from ~\cite{vondrick2016anticipating} as we are predicting the future motion using dynamic images. We make use reconstruction loss, class information loss, and expected future appearance as a guide to predict future motion images. As our generated dynamic images are trained for action anticipation, they are class specific and different from original dynamic images~\cite{bilen2016dynamic}. As demonstrated, our generated dynamic images are more effective than original dynamic images for action anticipation task. Gao~\etal~\cite{gao2017red} propose to generate future appearance using LSTM autoencoder to anticipate actions using both regression loss and classification loss. We argue that predicting future appearance representation is a complex task. We believe that action anticipation can benefit from motion prediction more than challenging appearance prediction.

Predicting the future content has been explored on other related problems in other domains of computer vision. Some of the work focuses on predicting (or forecasting) the future trajectories of pedestrians \cite{kitani2012activity} or predicting motion from still images \cite{kitani2012activity,pellegrini2009you}. However, we are the first to show the effectiveness of predicting good motion representations for early action anticipation.

%% file: method.tex
\begin{figure*}
    \centering
  \includegraphics[width=0.90\textwidth]{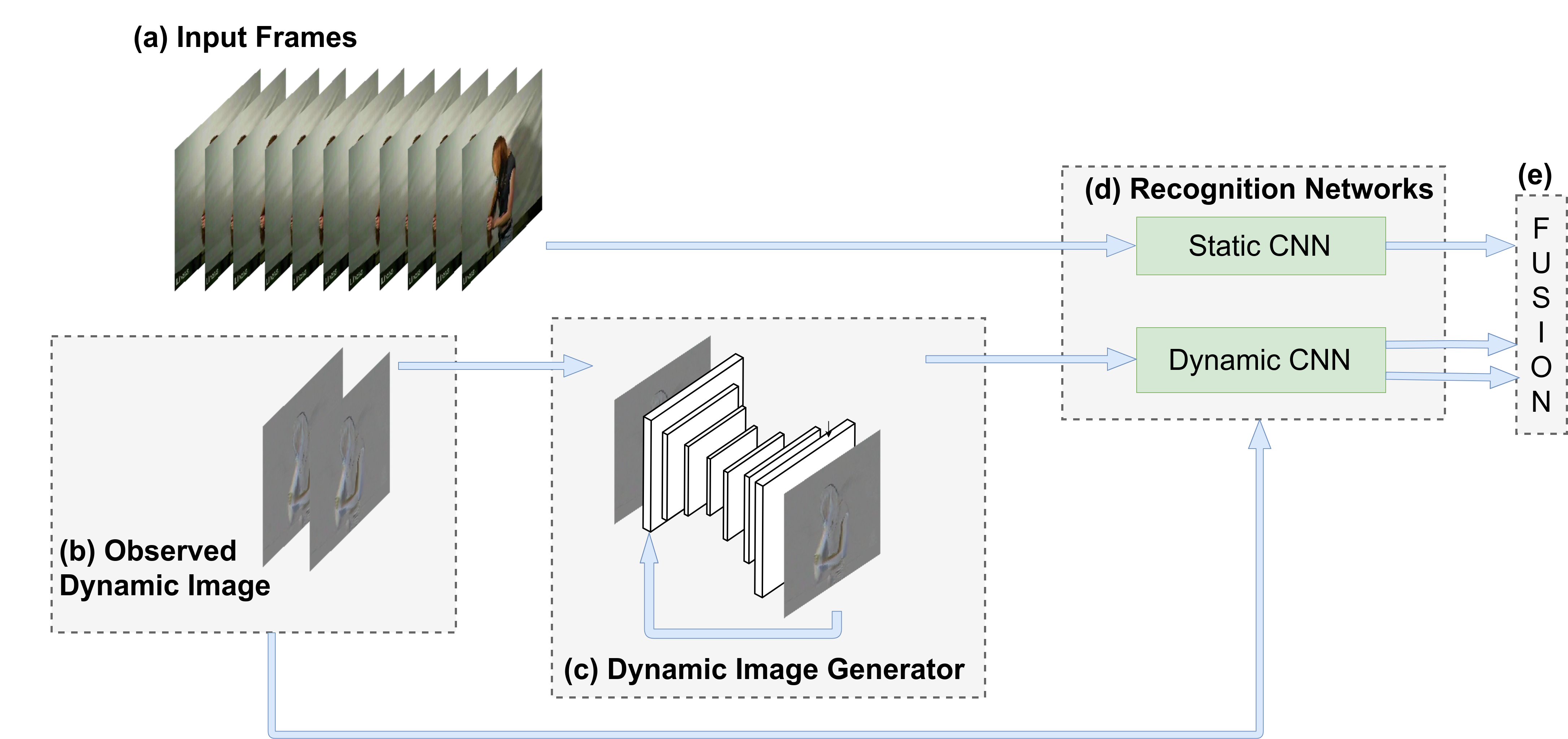}
    \caption{\textbf{Overview of our approach}. We receive as an input a sequence of RGB video frames \textbf{(a)}. Then we use RGB images with windows size $T$ to compute the Dynamic Images for seen part of the video \textbf{(b)}. The last dynamic image of the seen part is used to feed our dynamic image generator and generate $\hat{D}_{t+1}$ \textbf{(c)}. Next, we feed {\em Dynamic} CNN with observed dynamic images and generated dynamic images and {\em Static} CNN with RGB images \textbf{(d)}. Finally, we fusion all the outputs of our recognition networks \textbf{(e)}.} 
  \label{fig:overview}
\end{figure*}

The objective of our work is to recognise human actions as early as possible from a video sequence depicting human action.
We present a method that hallucinates future motion from a partially observed human action sequence (RGB video clip).
Then we process these hallucinated future motion representations to make future action predictions a.k.a. action anticipation.
Our motion representation is based on dynamic images~\cite{bilen2016dynamic,Bilen2017}.
Dynamic images model dynamic information of a short video clip and summarise motion information to a single frame.
We present a method to hallucinate future dynamic images using a convolutional autoencoder neural network. 
We process generated dynamic images to predict future human actions using a CNN named \emph{dynamic CNN}. 
To improve action recognition performance further, we use observed still image appearance information and process them with a \emph{static CNN}.
Furthermore, we make use of dynamic images created from observed RGB data and use the same dynamic CNN to make predictions.
Therefore, we make use of three kinds of predictions and fuse them to make the final prediction (see Fig.~\ref{fig:overview}).
In the following section, we present some background about dynamic images~\ref{sec.background} and then we present our dynamic image generation model in section~\ref{sec.model}. Then we discuss loss functions in section~\ref{sec.digen} and how to train our model in section~\ref{sec.mtl}.
\subsection{Background}
\label{sec.background}
Dynamic images~\cite{bilen2016dynamic,Bilen2017} are a compact motion representation of videos which is useful for human action recognition. 
They summarise the temporal evolution of a short video clip (\eg 10 frames) to a single still RGB image. 
Dynamic images are constructed using the rank pooling~\cite{FernandoGMGT15}. 
Rank pooling represents a video as a parameters of a linear ranking function that is able to chronologically order the elements of a sequence $\left<I_1, ..., I_T\right>$.
Precisely, let $\psi(I_t) \in \mathbb{R}^{d}$ be a feature vector extracted from each individual frame in the video and $V_t = \frac{1}{t} \sum_{\tau=1}^t \psi(I_\tau)$ be the average of these features up to time $t$. 
The ranking function $S(t|\mathbf{d})$ predicts a ranking score for each frame at time $t$ denoted by $S(t|\mathbf{d}) = \langle \mathbf{d}, V_t\rangle$, where $\mathbf{d} \in \mathbb{R}^d$ is the parameter of the linear ranking function~\cite{FernandoGMGT15}.
The parameter set $\mathbf{d}$ is learned so that the score reflect the rank of each frame.
Therefore, the ranking score for later frame at time $q$ ($q>t$) is associated with a larger score, \ie $S(q|\mathbf{d}) > S(t|\mathbf{d})$.
Learning $\mathbf{d}$ is posed as a convex optimisation problem using the RankSVM~\cite{smola2004tutorial} formulation given as equation~\ref{eq:rank}.
\begin{equation}
 \begin{multlined}
\mathbf{d}^*  = \rho(I_1, ... , I_t ; \psi) = \underset{d}{\mathrm{argmin}}~ E(\mathbf{d}), \\
E(\mathbf{d}) = \frac{\lambda}{2}||\mathbf{d}||^2 + 
\frac{2}{T(T-1)} \times \displaystyle \sum_{q>t} \max\{0, 1 - S(q|\mathbf{d}) + S(t|\mathbf{d})\}.
\end{multlined}
\label{eq:rank}
\end{equation}
Optimising equation~\ref{eq:rank} defines a function $\rho(I_1 , . . . , I_T ; \psi)$ that maps a video sequence of length $T$ to a single vector denoted by $\mathbf{d^∗}$. 
Since this parameter vector contains enough information to rank all frames in the video clip, it aggregates temporal information from all frames. 
Therefore, it  can be used as a video motion descriptor or a temporal descriptor.

When one applies this technique directly on RGB image pixels, the resulting $\mathbf{d^*}$ is known as the \emph{dynamic image}. 
The output $\mathbf{d^*}$ has same dimensions as input images.
Resulting dynamic image $\mathbf{d^*}$ summarises the temporal information of the RGB video sequence. 
Bilen~\etal~\cite{bilen2016dynamic} present an approximation to rank pooling which is faster.
This approximate rank pooling is essential for our method to hallucinate future dynamic images.
%
%
Bilen~\etal~\cite{bilen2016dynamic} proved that $\mathbf{d^*}$ can be expressed by the following equation~\ref{eq:di}.
\begin{equation}
 \mathbf{d^*} = \sum_{t=1}^T \alpha_t I_t.
 \label{eq:di}
\end{equation}
The coefficients $\alpha_t$ are given by
%
 $\alpha_t = 2(T - t + 1) - (T+1)(H_{T} - H_{t-1})$
%
where $H_t = \sum_{i=1}^t 1/t$  is the $t$-th Harmonic number and $H_0=0$. 
We construct dynamic images using approximated rank pooling by taking a weighted sum of input image sequence where weights are given by predefined coefficients $\alpha$.

\subsection{Future motion prediction model}
\label{sec.model}
Given a collection of videos $X$ with corresponding human action class labels $Y$, our aim is to predict the human action label as early as possible.

Each video $X_i \in X$ is a sequence of frames $X_i= \left< I_1, I_2, \cdots, I_n \right> $ of variable length $n$. 
We process each sequence of RGB frames to obtain a sequence of dynamic images using equation~\ref{eq:di}. 
Instead of summarising the entire video with a single dynamic image, we propose to generate multiple dynamic images from a single video sequence using a fixed window size of length $T$. 
Therefore, each dynamic image is created using $T$ consecutive frames. 
We process each training video $X_i$ and obtain a sequence of dynamic images $\left< D_1, D_2, \cdots, D_{n} \right> $. 
Our objective is to train a model that is able to predict the future dynamic image $D_{t+k}$ given the current dynamic images up to time $t$ \ie $\left< D_1, D_2, \cdots, D_t \right>$. 
Therefore, we aim to model the following conditional probability distribution using a parametric model
\begin{equation}
P(D_{t+k} | \left< D_1, D_2, \cdots, D_t \right>; \Theta)
\label{eq.prb.model}
\end{equation} 
where $\Theta$ are the parameters of our generative model ($k\ge1$). 
We simplify this probabilistic model using the Markov assumption, hence now $k=1$ and condition only on the previous dynamic image $D_t$. Then our model simplifies to following equation~\ref{eq.prb.model.simple}.
\begin{equation}
P(D_{t+1} | D_t ; \Theta)
\label{eq.prb.model.simple}
\end{equation} 
The model in equation~\ref{eq.prb.model.simple} simplifies the training process. 
Furthermore, it may be possible to take advantage of different kinds of neural machine to implement the model in equation~\ref{eq.prb.model.simple} such as autoencoders \cite{baldi2012autoencoders}, variational conditional autoencoders \cite{kingma2014semi,sohn2015learning} and conditional generative adversarial networks \cite{mirza2014conditional}.

Now the challenge is to find a good neural technique and loss function to train such a model. 
We use a denoising convolutional autoencoder to hallucinate future dynamic images given the current ones.
Our convolutional autoencoder receives a dynamic image at time $t$ and outputs a dynamic image for next time step $t+1$. 
In practice, dynamic images up to time $t$ is observed, and we recursively generate dynamic images for time $t+1, \cdots, t+k$ using Markov assumption.
Although we use a denoising convolutional autoencoder, our idea can also be implemented with other generative models.
The autoencoder we use has 4 convolution stages. Each convolution has kernels of size $5 \times 5$ with a stride of $2$ and the number of features maps for the convolution layers are set to 64, 128, 256, and 512 respectively. Then the deconvolution is the inverted mirror of the encoding network (see Fig~\ref{fig:overview}), which is inspired by the architecture used in DCGAN \cite{RadfordMC15}.
Next, we discuss suitable loss functions for training the autoencoder.
\subsection{Loss functions for training the autoencoder}
\label{sec.digen}

First, we propose make use of reconstructive loss coined \emph{Dynamic Loss} to reduce the $\mathcal{L}_2$ distance between predicted dynamic image $\hat{D}_{t+1}$ and the ground truth dynamic image obtained from the training data $D_{t+1}$ as shown in equation~\ref{eq:dl}.
\begin{equation}
 \mathcal{L}_{DL} = ||\hat{D}_{t+1} - D_{t+1}||_2
 \label{eq:dl}
\end{equation}
Even though this loss function helps us to generate expected future dynamic image, it does not guarantee that the generated dynamic image is discriminative for action anticipation. 
Indeed, we would like to generate a dynamic image that contains more action class information.
Therefore, we propose to explore the teacher-student networks~\cite{hinton2015distilling} to teach the autoencoder to produce dynamic images that would be useful for action anticipation.
First, we train a teacher CNN which takes dynamic images as input and produces the action category label. Let us denote this teacher CNN by $f(D_i;\Theta_{cnn})$ where it takes dynamic image $D_i$ and produces the corresponding class label vector $\hat{y_i}$. 
This teacher CNN that takes dynamic images as input and outputs labels is called \emph{Dynamic CNN} (see Fig~\ref{fig:overview}).
This teacher CNN is trained with cross-entropy loss~\cite{szegedy2017inception}.
Let us denote our generator network as $g(D_t;\Theta)\rightarrow D_{t+1}$. We would like to take advantage of the teacher network $f(;\Theta_{cnn})$ to guide the student generator $g(D_t;\Theta)$ to produce future dynamic images that are useful for classification. Given a collection of current and future dynamic images with labels, we train the generator with the cross-entropy loss as follows:
\begin{equation}
 \mathcal{L}_{CL} = - \sum_t y_i \log f(g(D_t;\Theta);\Theta_{cnn})
\end{equation}
where we fix the CNN parameter $\Theta_{cnn}$. Obviously, we make the assumption that CNN $f(D_i;\Theta_{cnn})$ is well trained and has good generalisation capacity. 
We call this loss as the \emph{classification loss} which is denoted by $\mathcal{L}_{CL}$. 
In theory, compared to original dynamic images~\cite{bilen2016dynamic,Bilen2017}, our generated dynamic images are class specific and therefore discriminative.

Motion and appearance are related.
Optical flow depends on the appearance of two consecutive frames.
Dynamic images depends on the evolution of appearance of several consecutive frames.
Therefore, it is important verify that generated future motion actually adhere to future expected appearance.
Another advantage of using dynamic images to generate future motion is the ability exploit this property explicitly.
We make use of future expected appearance to guide the generator network to produce accurate dynamic images.
Let us explain what we mean by this. 
When we generate future dynamic image $D_{t+1}$, as demonstrated in equation \ref{eq:rgb}, implicitly we also recover the future RGB frame $I_{t+1}$. 
Using this equation~\ref{eq:rgb}, we propose so-called \emph{static loss} (SL) (equation~\ref{eq:sl}) that consists of computing the $\mathcal{L}2$ loss between the generated RGB image $\hat{I}_{t+1}$ and real expected image $I_{t+1}$.
\begin{align}
\label{eq:rgb}
D_{t+1} &= \displaystyle \sum_{i=1}^{T} \alpha_{i} I_{t+1+i} \\ \nonumber
D_{t+1} &= \alpha_{T}I_{T+t+1} \displaystyle \sum_{i=1}^{T-1} \alpha_{i} I_{t+1+i} \\ \nonumber
I_{T+t+1} &= \frac{D_{t+1} - \displaystyle \sum_{i=1}^{T-1} \alpha_{i} I_{t+1+i}}{\alpha_{T}}
\end{align}
The applicability of static loss does not limit only to matching the future expected appearance, but also we guide the autoencoder model $g(;\Theta)$ to use all implicitly generated RGB frames from $\hat{I}_{t+2}$ to $\hat{I}_{T+t+1}$ making future dynamic image better by modeling the evolution of appearance of static images. Indeed, this is a better loss function than simply taking the dynamic loss as in equation~\ref{eq:dl}.
\begin{equation}
 \mathcal{L}_{SL} = ||\hat{I}_{T+t+1} - I_{T+t+1}||_2
 \label{eq:sl}
\end{equation}
\subsection{Multitask learning}
\label{sec.mtl}
We train our autoencoder with multiple losses, the static loss ($\mathcal{L}_{SL}$), the dynamic loss ($\mathcal{L}_{DL}$) and the classification loss ($\mathcal{L}_{CL}$). By doing so, we aim to generate dynamic images that are good for the classification, as well as representative of future motion. 
With the intention to enforce all these requirements, we propose to train our autoencoder with batch wise multitask manner. 
Overall, one might write down the global loss function $\mathcal{L} =  \lambda_{sl} \mathcal{L}_{SL} + \lambda_{dl} \mathcal{L}_{DL} + \lambda_{cl} \mathcal{L}_{CL}$. 
However, instead of finding good scalar weights $\lambda_{sl}, \lambda_{dl},$ and $\lambda_{cl}$, we propose to divide each batch into three sub-batches, and optimise each loss using only one of those sub batches. Therefore, during each batch, we optimise all losses with different sets of data. 
We found this strategy leads to better generalisation than optimising a linear combination of losses.

\subsection{Inference}
During inference, we receive RGB frames from a video sequence as input. Using those RGB frames, we compute \textit{dynamic images} following equation \ref{eq:di} with a window size length $T=10$. In the case that the amount of frames is less that what is needed to compute the dynamic image i.e. 10\% of the video is observed, we compute the dynamic image with the available frames according to equation~\ref{eq:di}.
We use the last dynamic image ($D_t$) to predict the following dynamic image ($\hat{D}_{t+1}$). 
We repeat this process to generate $k$ number of future dynamic images using Markov assumption.
We process each observed RGB frame, observed dynamic images and generated dynamic images by respective static and dynamic CNNs that are trained to make predictions (see Fig.~\ref{fig:overview}). 
Then, we obtain a score vector for each RGB frame, dynamic image and generated dynamic image.
We sum them together and use temporal average pooling to make the final prediction. 

%% file: experiments.tex
In this section, we perform a series of experiments to evaluate our action anticipation method. First, we present results for action recognition using the \emph{static CNN} and the \emph{dynamic CNN} in section~\ref{sec.exp.action}. Then, we evaluate the impact of different loss functions for generating future dynamic images in section~\ref{sec.exp.loss}. After that in section~\ref{sec.act.eva}, we compare our method with state-of-the-art techniques for action anticipation. Finally, we present some other additional experiments to further analyse our model in sections~\ref{sec.exp.further}.

\textbf{Datasets} We test our method using three popular datasets for human action analysis JHMDB~\cite{jhuang2013towards}, UT-Interaction~\cite{UTInteractionData} and  UCF101-24~\cite{soomro2012ucf101}, which have been used for action anticipation in recent prior works~\cite{aliakbarian2017encouraging,soomro2016predicting,Singh2017}. 

\textbf{JHMDB} dataset is a subset of the challenging HMDB51 dataset \cite{kuehne2011hmdb}. JHMDB is created by keeping action classes that involve a single person action. Videos have been collected from different sources such as movies and the world-wide-web. JHMDB dataset consists of 928 videos and 21 action classes. Each video contains one human action which usually starts at the beginning of the video. Following the recent literature for action anticipation~\cite{aliakbarian2017encouraging}, we report the average accuracy over the three splits and report results for so called \emph{earliest} setup. For earliest recognition, action recognition performance is measured only after observing 20\% of the video. To further understand our method, we also report recognition performance w.r.t. time (as a percentage).
\textbf{UT-Interaction} dataset (UTI) contains 20 video sequences where the average length of a video is around 1 minute. These videos contain complete executions of 6 human interaction classes: shake-hands, point, hug, push, kick and punch. Each video contains at least one execution of an interaction, and up to a maximum of 8 interactions. There are more than 15 different participants with different clothing. The videos are recorded with 30fps and with a resolution of 720 x 480 which we resize to 320 x 240. To evaluate all methods, we use recommended 10-fold leave-one-out cross-validation per set and report the mean performance over all sets. 
\textbf{UCF101-24} dataset is a subset of the challenging UCF101 dataset. This subset of 24 classes contains spatio-temporal localisation annotation. It has been constructed for THUMOS-2013 challenge\footnote{http://crcv.ucf.edu/ICCV13-Action-Workshop/download.html}. On average there are 1.5 action instances per video, each instance cover approximately 70\% of the duration of video. We report the action-anticipation accuracy for set 1, as has been done previously in \cite{Singh2017}. 


\subsection{Training of {\em Static} and {\em Dynamic} CNNs.}
\label{sec.exp.action}
\begin{table}[t]
  \centering
  \begin{tabular}{lc@{\hskip 0.1in}cc}
  \hline
  & \textbf{JHMDB} & \textbf{UT-Interaction} \\ \hline
Static CNN & 55.0\% & 70.9\%  \\
Dynamic CNN & 54.1\% & 71.8\% \\
\hline
  \end{tabular}
  \caption{Action recognition performance using dynamic and RGB images over JHMDB and UT-Interaction datasets. Action recognition performance is measured at frame level.}
  \label{tab:compl}
\end{table}
In this section, we explain how we train our \emph{static} and \emph{dynamic} CNNs (see Fig.~\ref{fig:overview}).
Similar to~\cite{bilen2016dynamic,Bilen2017}, we train a \emph{Static CNN} for RGB frame-based video action recognition and a \emph{Dynamic CNN} for dynamic image-based action recognition. In all our experiments, each dynamic image is constructed using 10 RGB frames (T=10). We use different data augmentation techniques to reduce the effect of over-fitting. Images are randomly flipped horizontally, rotated by a random amount in a range of -20 to 20 degrees, horizontally shifted in a range of -64 to 64 pixels, vertically shifted in a range of -48 to 48 pixels, sheared in a range of 10 degrees counter-clockwise, zoomed in a range of 0.8 to 1.2 and shifted channels in a range of 0.3. We make use of pre-trained Inception Resnet V2 \cite{szegedy2017inception} to fine-tune both \emph{Static CNN} and the \emph{Dynamic CNN} using a learning rate of 0.0001. We use a batch size of 32 and a weight decay of 0.00004. We use ADAM \cite{kingma2015adam} optimizer to train these networks using epsilon of 0.1 and beta 0.5.
Action recognition performance using these CNNs for JHMDB and UTI datasets are reported in Table \ref{tab:compl}. Note that the action recognition performance in Table \ref{tab:compl} is only at frame level (not video level).
We use these trained {\em Static} and {\em Dynamic} CNNs in the generation of future motion representation, dynamic images, and action anticipation tasks.

\subsection{Impact of loss functions.}
\label{sec.exp.loss}
In this section we investigate the effectiveness of each loss function, explained in section~\ref{sec.digen}, in the generation process of future dynamic images. We evaluate the quality of the generated dynamic images in a {\em quantitative} evaluation. Using the dynamic CNN to report action recognition performance over generated dynamic images. 

We perform this experiment constructing a sequence of dynamic images using equation~\ref{eq:di} for each test video in the dataset. Then for each test dynamic image, we generate the future dynamic image using our convolutional autoencoder. Therefore, the number of generated dynamic images is almost equal to real testing dynamic images. Then we use our dynamic CNN (which has been pretrained in previous section) to evaluate the action recognition performance of generated dynamic images (\textbf{DIg}). Using this approach we can evaluate the impact of several loss functions in the generation of dynamic images. 

We use the first split of JHMDB and the first set of UTI to perform this experiment. We make use of the three proposed losses in section~\ref{sec.digen}: dynamic-loss ($\mathcal{L}_{DL}$), class-based loss ($\mathcal{L}_{CL}$) and static-loss ($\mathcal{L}_{SL}$) to train our autoencoder. We train the convolutional autoencoder using ADAM solver with a batch size of 32, a learning rate of 0.0001. We train our model for 30 epochs using the same augmentation process used in section~\ref{sec.exp.action}. 

We use the generalisation performance of {\em real dynamic images} from Table~\ref{tab:compl} as a reference to estimate the quality of generated dynamic images. Since, we measure the performance of generated dynamic images in the same way.


\begin{table}[t]
    \centering
\begin{tabular}{lc@{\hskip 0.1in}c}
\hline                                                         & \textbf{JHMDB-21} &  \textbf{UT-Interaction}\\ \hline
$\mathcal{L}_{DL}$                                       & 42.8\%              &  64.3\%\\
$\mathcal{L}_{SL}$                                       & 49.5\%              &  64.2\%\\
$\mathcal{L}_{DL} + \mathcal{L}_{SL}$                    & 53.4\%              &  66.5\%\\
$\mathcal{L}_{DL} + \mathcal{L}_{CL}$                    & 52.5\%              &  64.5\%\\
$\mathcal{L}_{DL} + \mathcal{L}_{SL} + \mathcal{L}_{CL}$ & 54.0\%              &  68.4\%\\ \hline
\end{tabular}
  \caption{Results of using multitask learning to generate future dynamic images.}
  \label{tab:digen}
\end{table}

As can be seen in Table~\ref{tab:digen}, a combination of $\mathcal{L}_{DL}$, $\mathcal{L}_{CL}$ and $\mathcal{L}_{SL}$ gives excellent recognition performance of 54.0\% for the generated dynamic images which is very close to the model performance of single dynamic CNN 54.1\% in the case of JHMDB dataset. Indicating that our generative model along with loss functions are capable of generating representative and useful future dynamic images.
A similar trend can be seen for UTI dataset. 
Notice that the $\mathcal{L}_{DL}$ and $\mathcal{L}_{SL}$ already produce good recognition performance on JHMDB and UTI datasets, which suggest that those losses can generated images that understand the human motion. However, those generated images are not class specific.
We conclude that convolutional autoencoder model trained with three losses is able to generate robust future dynamic images. These generated dynamic images are effective in action recognition.
\subsection{Action anticipation}
\label{sec.act.eva}
\begin{figure}[t]
  \begin{subfigure}[b]{0.5\textwidth}
    \includegraphics[width=\textwidth]{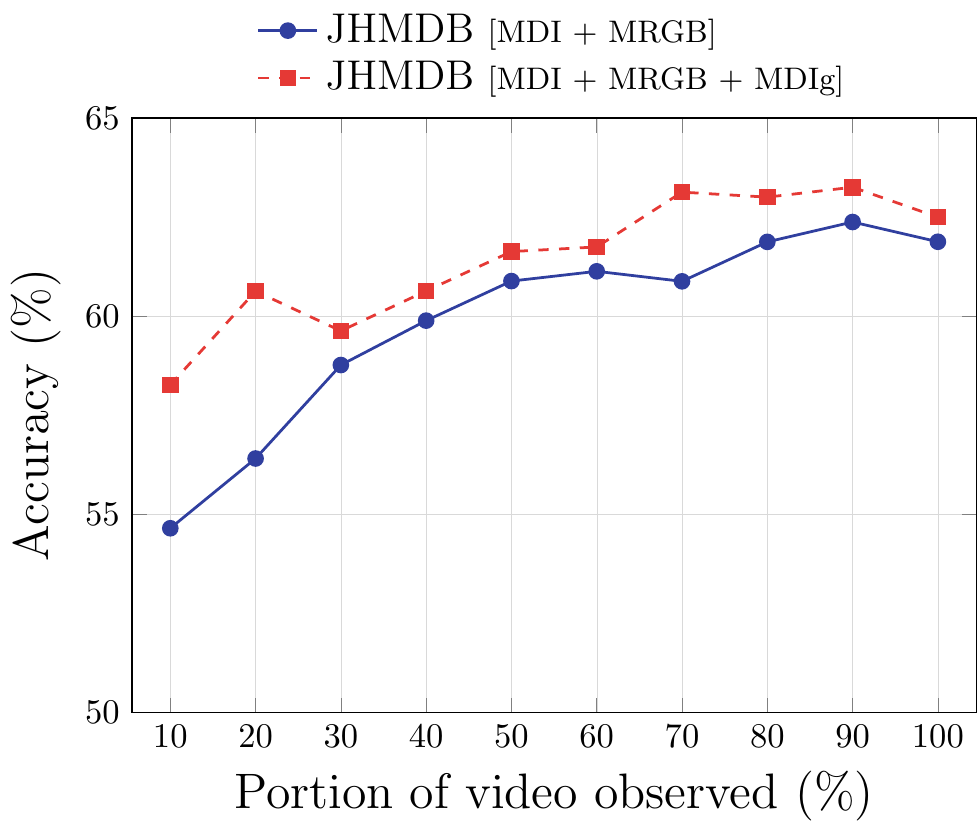}
  \end{subfigure}
  \hfill
  \begin{subfigure}[b]{0.5\textwidth}
    \includegraphics[width=\textwidth]{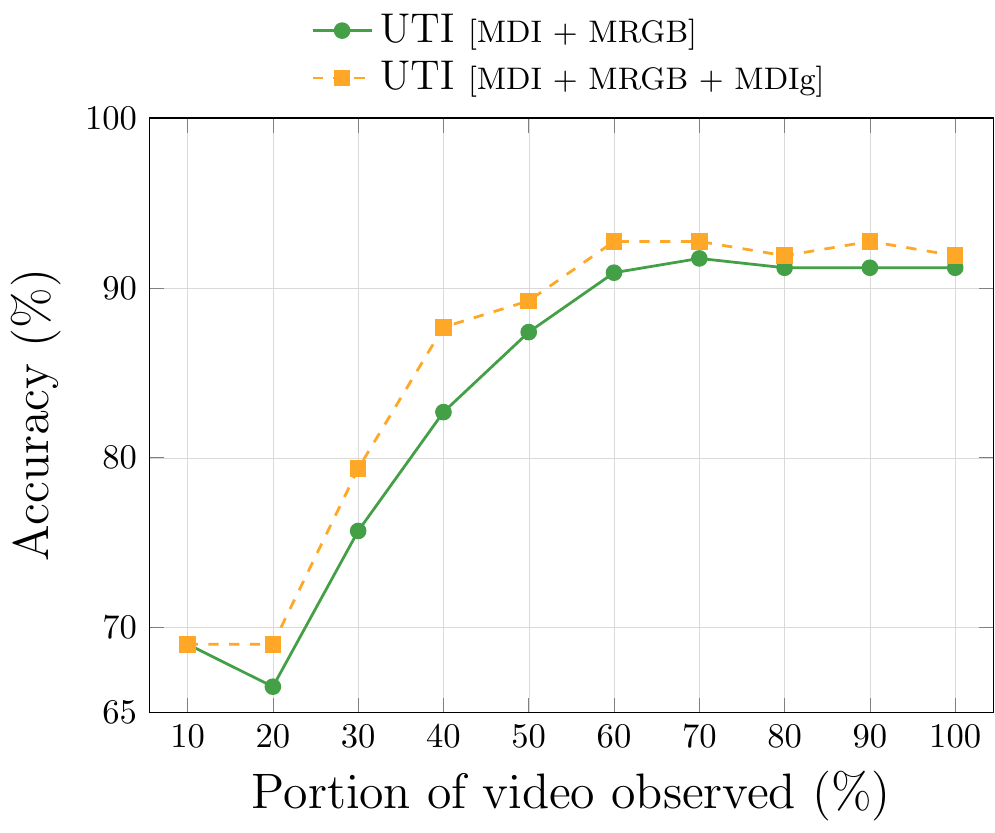}
  \end{subfigure}
  \caption{Action anticipation performance with respect to portion of the video observed on JHMDB {\em(left)} and UTI {\em (right)} datasets.}
  \label{fig:ratiovideo}
\end{figure}

Our action anticipation network consist of a {\em static} CNN and a {\em dynamic} CNN (see Fig~\ref{fig:overview}). Our action anticipation baseline uses observed multiple RGB frames and multiple dynamic images similar to~\cite{bilen2016dynamic}. In addition to that our method generates K number of future dynamic images and make use of them with dynamic CNN. Action anticipation performance is evaluated at different time steps after observing fraction of the video (\ie, 10$\%$, 20$\%$, $\cdots$, 100$\%$ of the video). Results are shown in Figure~\ref{fig:ratiovideo}. We can see that the most significant improvement for JHMDB is obtained at 20\% which is an enhancement of \textbf{5.1\%} with respect to the baseline. In the case of UTI dataset, the most significant improvement is obtained at 40\% of the video observed with a performance enhancement of \textbf{5.0\%} with respect to the baseline. Moreover, the less significant improvement are obtained when the video observation approaches the 100\% with a 0.62\% and 0.71\% of improvement with respect to the baseline on JHMDB and UTI dataset respectively.


Another standard practice is to report the action anticipation performance using {\em earliest} and {\em latest} prediction accuracies as done in~\cite{ryoo2011human,aliakbarian2017encouraging}. Although, there is no agreement of what is the proportion of frames used in earliest configuration through different datasets. We make use of the proportion that has been employed by baselines (20\% and 50\% of the video for JHMDB and UTI respectively). Therefore, following~\cite{aliakbarian2017encouraging} we report results in Table \ref{tab:res:jhmdb21}, Table~\ref{tab:res:UTinter} for JHMDB and UTI datasets respectively.
We outperform other methods that rely on additional information, such as optical flow \cite{ma2016learning,soomro2016online,soomro2016predicting} and Fisher vector features based on improved Dense Trajectories \cite{soomro2016online}. Our approach outperforms the state-of-the-art by \textbf{4.0\%} on JHMDB and by \textbf{5.2\%} on UTI datasets in the  earliest configuration. Finally, we report results on UCF101-24 dataset for action anticipation. For this dataset, we use 10\% of the video to predict the action class in the earliest configuration. As we can see in Table \ref{tab:res:ucf101}, We outperform previous method \cite{Singh2017} by \textbf{5.1\%} on the earliest configuration. A more detailed analysis using UCF101-24 dataset is provided on the supplementary material.

\begin{table}[t]
    \centering
    \scalebox{0.9}{
    \begin{minipage}[b]{0.5\textwidth}
        \centering        
        \begin{tabular}{lcc}
            \\ \hline
            Method                                           & Earliest & Latest \\ \hline
            DP-SVM \cite{soomro2016online}                 & 5\%      & 46\%      \\
            S-SVM \cite{soomro2016online}                  & 5\%      & 43\%      \\
            Where/What \cite{soomro2016predicting}         & 12\%     & 43\%      \\
            Context-Aware+Loss of \cite{jain2016recurrent} & 28\%     & 43\%      \\
            Ranking Loss \cite{ma2016learning}             & 29\%     & 43\%      \\
            Context-Aware+Loss of \cite{ma2016learning}    & 33\%     & 39\%      \\
            E-LSTM \cite{aliakbarian2017encouraging}       & 55\%     & 58\%      \\ 
            ROAD \cite{Singh2017} & 57\% & 68\% \\ \hline
            Ours                                          & \textbf{61\%}     & 63\%     \\
            \hline
        \end{tabular}
        \caption{Comparison of action anticipation methods on \textbf{JHMDB} dataset. 20\% of video is observed at \emph{Earliest}.}
        \label{tab:res:jhmdb21}
      \begin{tabular}{lll}
      \hline
              & Earliest   & Latest   \\ \hline
        ROAD (RTF) \cite{Singh2017} & 81.7\% & 83.9\% \\
        ROAD (AF)  \cite{Singh2017} & 84.2\% & 85.5\% \\ \hline
        Ours  & 89.3\% & 90.2\% \\ \hline
      \end{tabular}
      \caption{Comparison of action anticipation methods on \textbf{UCF101-24} dataset. 10\% of video is observed at Earliest.}
      \label{tab:res:ucf101}
    \end{minipage}%
    }
    \qquad
    \scalebox{0.9}{
    \begin{minipage}[b]{0.5\textwidth}
        \centering
        \begin{tabular}{lll}
            \\ \hline
            Method                & Earliest & Latest \\ \hline
            S-SVN \cite{soomro2016online} & 11.0\%   & 13.4\% \\
            DP-SVM \cite{soomro2016online} & 13.0\%   & 14.6\% \\
            CuboidBayes \cite{ryoo2011human} & 25.0\%   & 71.7\% \\
            CuboidSVM \cite{ryoo2010overview} & 31.7\%   & 85.0\% \\
            Context-Aware+Loss of \cite{jain2016recurrent}& 45.0\%   & 65.0\% \\
            Context-Aware+Loss of \cite{ma2016learning}& 48.0\%   & 60.0\% \\ 
            BP\_SVM \cite{laviers2009improving} & 65.0\%   & 83.3\% \\
            I-BoW \cite{ryoo2011human} & 65.0\%   & 81.7\% \\
            D-BoW \cite{ryoo2011human} & 70.0\%   & 85.0\% \\
            E-LSTM \cite{aliakbarian2017encouraging} & 84.0\%   & 90.0\% \\             
            \hline
            Ours & 89.2\% & 91.9\% \\ \hline
        \end{tabular}
        \caption{Comparison of action anticipation methods using \textbf{UTI} dataset. 50\% of the video is observed at \emph{Earliest}.}
        \label{tab:res:UTinter}
    \end{minipage}
    }
\end{table}

These experiments evidence the benefits of generating future motion information using our framework for action anticipation.
\subsection{Further exploration}
\label{sec.exp.further}

\begin{figure}[t]
\centering
\caption{Impact of generating more future dynamic images recursively on JHMDB \textit{(left)} and UTI \textit{(right)} datasets. K is the number of generated dynamic images based on observed RGB frames. K=0 means no dynamic image is generated.}
\label{tab:woworgbjhmdb}
\includegraphics[width=0.45\textwidth]{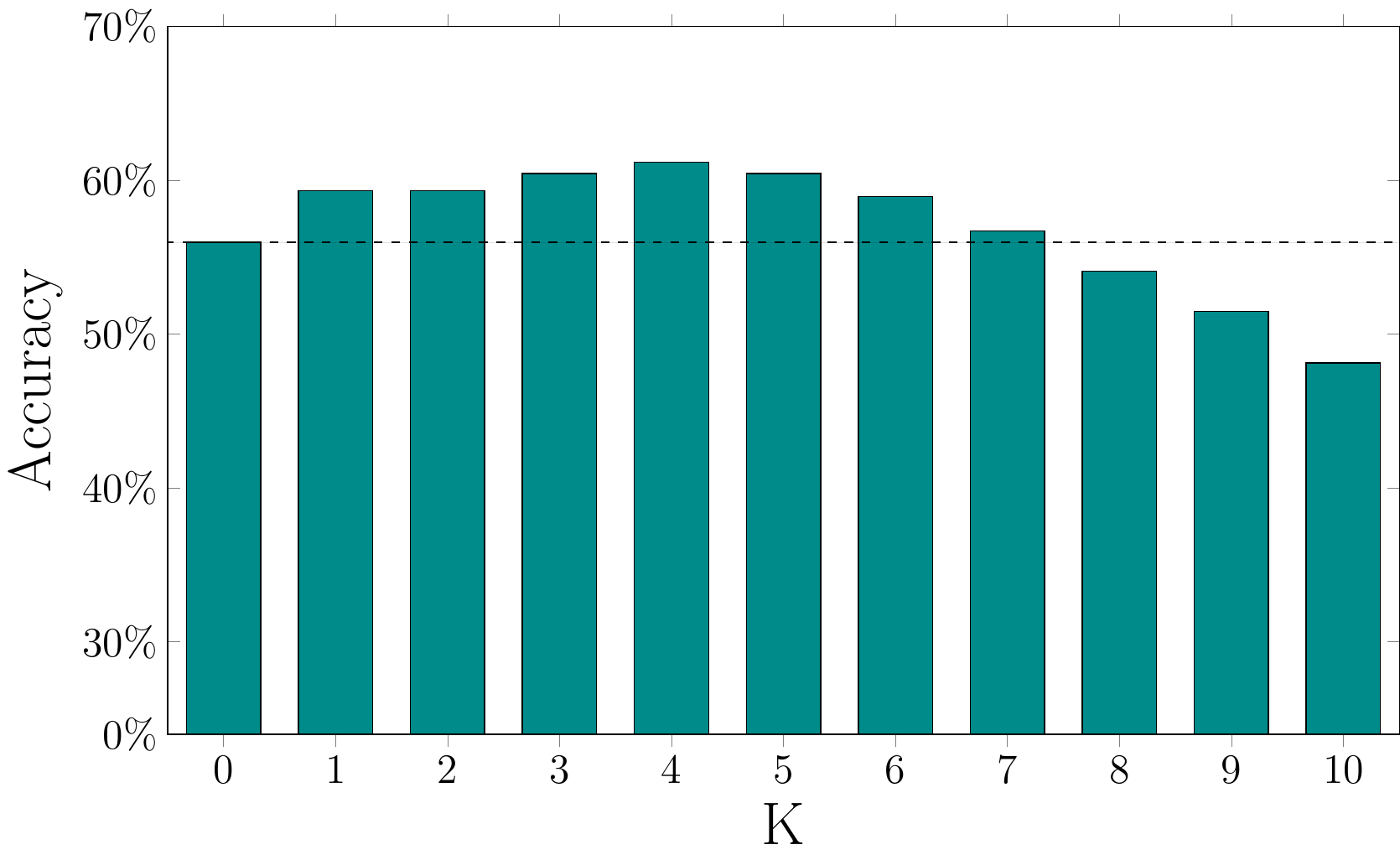}
\includegraphics[width=0.45\textwidth]{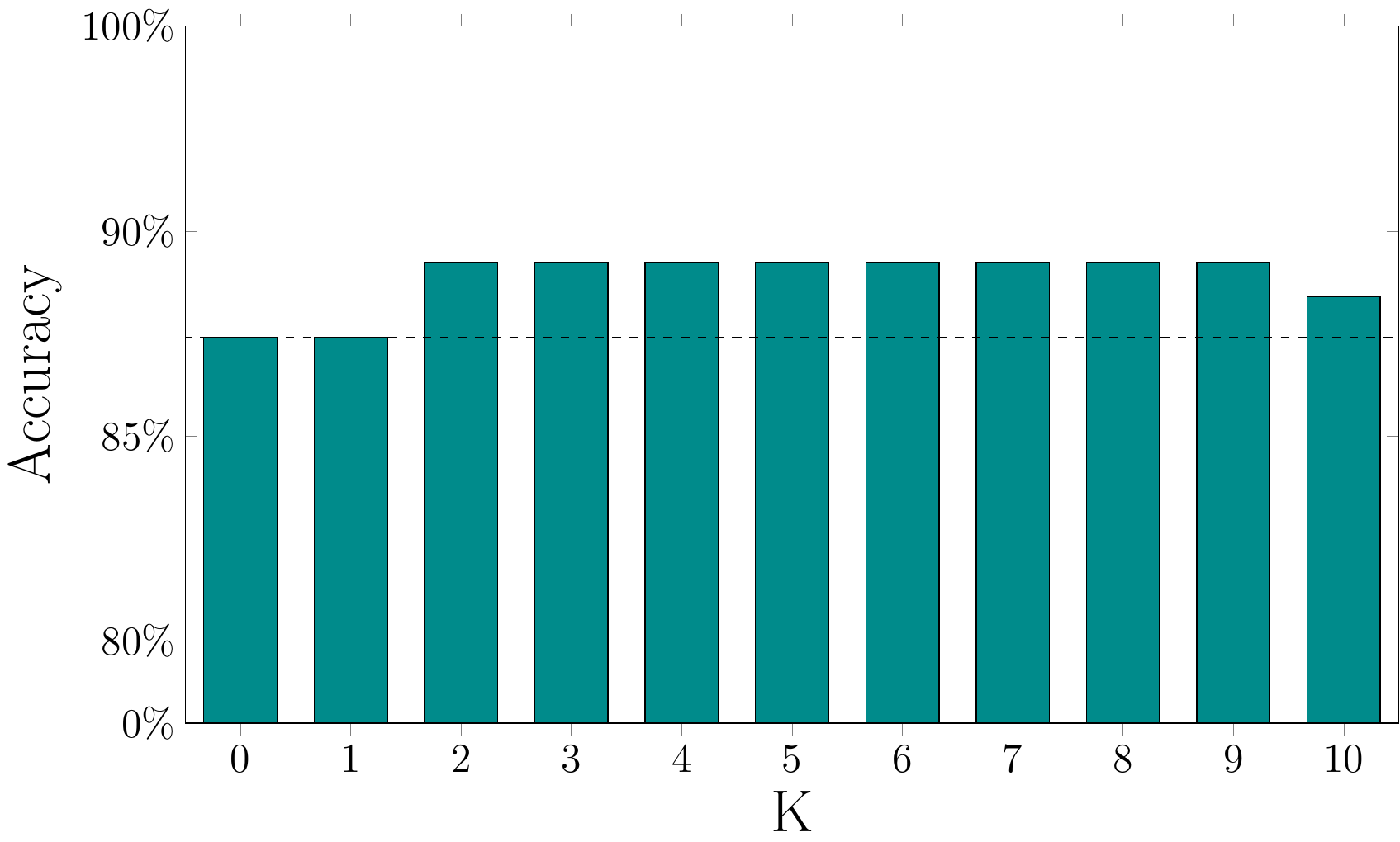}
\end{figure}

In Fig.~\ref{tab:woworgbjhmdb} we observe the influence of generating dynamic images recursively for earliest configuration in JHMDB and UTI datasets. 
We generate $K$ number of future dynamic images recursively using the very last true dynamic image. 
As it can be seen in Fig.~\ref{tab:woworgbjhmdb}, as we generate more dynamic images into the future, the prediction performance degrades due to the error propagation.
We report action recognition performance for each generated future dynamic image (\ie for the generated future dynamic image at $K$).
If we do not generate any dynamic image for the future, we obtain an action recognition performance of 55.9\%.
If we include generated dynamic images, we obtain a best of 61.0\% on JHMDB. 
A similar trend can be seen for UTI dataset, where without future dynamic image we obtain 87.4\% and after generation we obtain an action recognition performance of 89.2\%.
The influence of generating more future dynamic images is shown in Fig~\ref{tab:woworgbjhmdb}.

\begin{figure*}[t]
 \centering
 \caption{Visual comparison between generated dynamic image {\em (bottom)} and ground truth {\em (top)}. $K$ refers to how many iterations we apply in the generation of dynamic image.}
 \label{fig:visual}
 \includegraphics[width=0.9\textwidth]{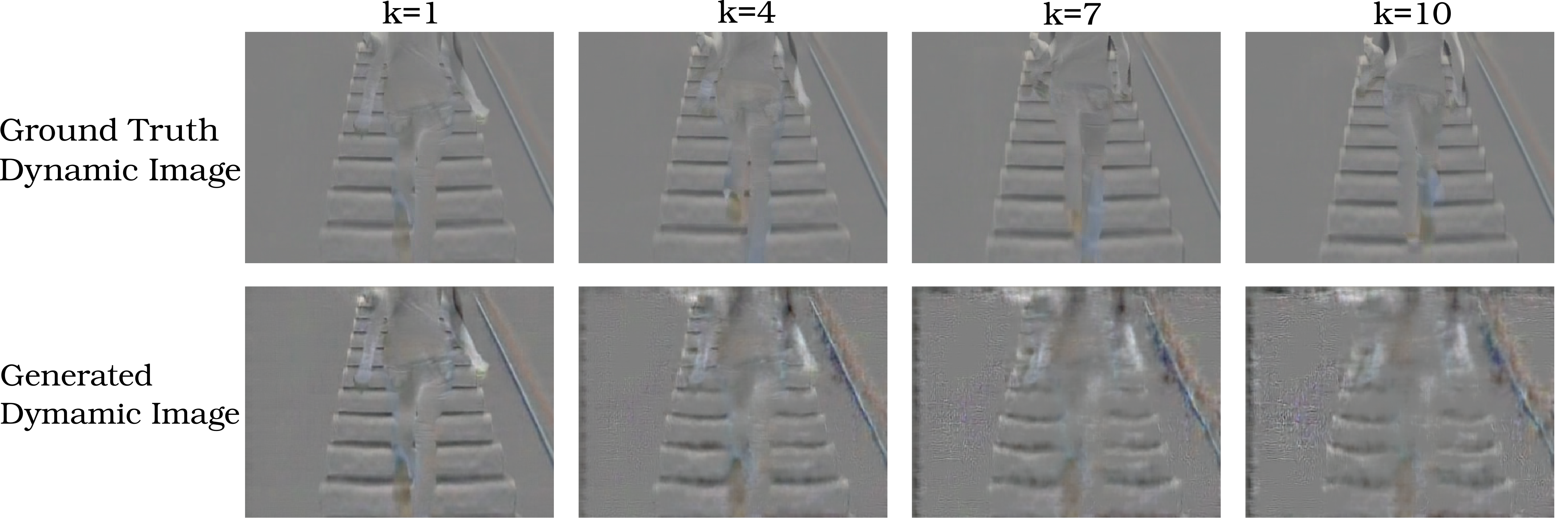}
\end{figure*}
Finally, we visually inspect the recursively generated dynamic images for $K$ equal to 1, 4, 7 and 10 in Fig.~\ref{fig:visual}. 
Although, we can use our model to generate quite accurate dynamic images, as we predict into the further, the generated dynamic images might contain some artifacts.

%% file: discussion.tex
In this paper, we demonstrate how to hallucinate future video motion representation for action anticipation. We propose several loss functions to train our generative model in a multitask scheme. Our experiments demonstrate the effectiveness of our loss functions to produce better future video representation for the task of action anticipation. Moreover, experiments show that made use of the hallucinated future video motion representations improves the action anticipation results of our powerful backbone network. With our simple approach we have outperformed the state-of-the-art in action anticipation in three important action anticipation benchmarks. In the future, we would like to incorporate additional sources of information to hallucinate other dynamics such as optical flow using the same framework. Furthermore, we would like to extend this method to predict dynamic images further into the future.

%% file: eccv2018submission.bbl
\begin{thebibliography}{10}

\bibitem{bilen2016dynamic}
Bilen, H., Fernando, B., Gavves, E., Vedaldi, A., Gould, S.:
\newblock Dynamic image networks for action recognition.
\newblock In: CVPR. (2016)

\bibitem{ma2016learning}
Ma, S., Sigal, L., Sclaroff, S.:
\newblock Learning activity progression in lstms for activity detection and
  early detection.
\newblock In: CVPR. (2016)

\bibitem{ryoo2011human}
Ryoo, M.S.:
\newblock Human activity prediction: Early recognition of ongoing activities
  from streaming videos.
\newblock In: ICCV. (2011)

\bibitem{aliakbarian2017encouraging}
Sadegh~Aliakbarian, M., Sadat~Saleh, F., Salzmann, M., Fernando, B., Petersson,
  L., Andersson, L.:
\newblock Encouraging lstms to anticipate actions very early.
\newblock ICCV (2017)

\bibitem{soomro2016online}
Soomro, K., Idrees, H., Shah, M.:
\newblock Online localization and prediction of actions and interactions.
\newblock arXiv:1612.01194 (2016)

\bibitem{soomro2016predicting}
Soomro, K., Idrees, H., Shah, M.:
\newblock Predicting the where and what of actors and actions through online
  action localization.
\newblock In: CVPR. (2016)

\bibitem{lan2014hierarchical}
Lan, T., Chen, T.C., Savarese, S.:
\newblock A hierarchical representation for future action prediction.
\newblock In: ECCV. (2014)

\bibitem{yu2012predicting}
Yu, G., Yuan, J., Liu, Z.:
\newblock Predicting human activities using spatio-temporal structure of
  interest points.
\newblock In: ACMMM. (2012)

\bibitem{Bilen2017}
Bilen, H., Fernando, B., Gavves, E., Vedaldi, A.:
\newblock Action recognition with dynamic image networks.
\newblock IEEE Transactions on Pattern Analysis and Machine Intelligence
  \textbf{PP}(99) (2017)  1--1

\bibitem{Simonyan2014}
Simonyan, K., Zisserman, A.:
\newblock Two-stream convolutional networks for action recognition in videos.
\newblock In: NIPS. (2014)

\bibitem{Ahad2012}
Ahad, M.A.R., Tan, J.K., Kim, H., Ishikawa, S.:
\newblock Motion history image: its variants and applications.
\newblock Machine Vision and Applications \textbf{23}(2) (2012)  255--281

\bibitem{Carreira2017}
Carreira, J., Zisserman, A.:
\newblock Quo vadis, action recognition? a new model and the kinetics dataset.
\newblock In: CVPR. (2017)

\bibitem{Yu2012}
Yu, G., Yuan, J., Liu, Z.:
\newblock Predicting human activities using spatio-temporal structure of
  interest points.
\newblock In: ACMMM. (2012)

\bibitem{Li2014}
Li, K., Fu, Y.:
\newblock Prediction of human activity by discovering temporal sequence
  patterns.
\newblock IEEE Transactions on Pattern Analysis and Machine Intelligence
  \textbf{36}(8) (2014)  1644--1657

\bibitem{kong2014discriminative}
Kong, Y., Kit, D., Fu, Y.:
\newblock A discriminative model with multiple temporal scales for action
  prediction.
\newblock In: ECCV. (2014)

\bibitem{vondrick2016anticipating}
Vondrick, C., Pirsiavash, H., Torralba, A.:
\newblock Anticipating visual representations from unlabeled video.
\newblock In: CVPR. (2016)

\bibitem{jain2016recurrent}
Jain, A., Singh, A., Koppula, H.S., Soh, S., Saxena, A.:
\newblock Recurrent neural networks for driver activity anticipation via
  sensory-fusion architecture.
\newblock In: ICRA. (2016)

\bibitem{Singh2017}
Singh, G., Saha, S., Sapienza, M., Torr, P.H.S., Cuzzolin, F.:
\newblock Online real-time multiple spatiotemporal action localisation and
  prediction.
\newblock In: ICCV. (2017)

\bibitem{gao2017red}
Gao, J., Yang, Z., Nevatia, R.:
\newblock Red: Reinforced encoder-decoder networks for action anticipation.
\newblock arXiv:1707.04818 (2017)

\bibitem{kitani2012activity}
Kitani, K.M., Ziebart, B.D., Bagnell, J.A., Hebert, M.:
\newblock Activity forecasting.
\newblock In: ECCV. (2012)

\bibitem{pellegrini2009you}
Pellegrini, S., Ess, A., Schindler, K., Van~Gool, L.:
\newblock You'll never walk alone: Modeling social behavior for multi-target
  tracking.
\newblock In: ICCV. (2009)

\bibitem{FernandoGMGT15}
Fernando, B., Gavves, E., Oramas, J., Ghodrati, A., Tuytelaars, T.:
\newblock Rank pooling for action recognition.
\newblock IEEE Transactions on Pattern Analysis and Machine Intelligence
  \textbf{39}(4) (2017)  773--787

\bibitem{smola2004tutorial}
Smola, A.J., Sch{\"o}lkopf, B.:
\newblock A tutorial on support vector regression.
\newblock Statistics and computing \textbf{14}(3) (2004)  199--222

\bibitem{baldi2012autoencoders}
Baldi, P.:
\newblock Autoencoders, unsupervised learning, and deep architectures.
\newblock In: ICML. (2012)

\bibitem{kingma2014semi}
Kingma, D.P., Mohamed, S., Rezende, D.J., Welling, M.:
\newblock Semi-supervised learning with deep generative models.
\newblock In: NIPS. (2014)

\bibitem{sohn2015learning}
Sohn, K., Lee, H., Yan, X.:
\newblock Learning structured output representation using deep conditional
  generative models.
\newblock In: NIPS. (2015)

\bibitem{mirza2014conditional}
Mirza, M., Osindero, S.:
\newblock Conditional generative adversarial nets.
\newblock arXiv:1411.1784 (2014)

\bibitem{RadfordMC15}
Radford, A., Metz, L., Chintala, S.:
\newblock Unsupervised representation learning with deep convolutional
  generative adversarial networks.
\newblock ICLR (2016)

\bibitem{hinton2015distilling}
Hinton, G., Vinyals, O., Dean, J.:
\newblock Distilling the knowledge in a neural network.
\newblock arXiv:1503.02531 (2015)

\bibitem{szegedy2017inception}
Szegedy, C., Ioffe, S., Vanhoucke, V., Alemi, A.A.:
\newblock Inception-v4, inception-resnet and the impact of residual connections
  on learning.
\newblock In: AAAI. (2017)

\bibitem{jhuang2013towards}
Jhuang, H., Gall, J., Zuffi, S., Schmid, C., Black, M.J.:
\newblock Towards understanding action recognition.
\newblock In: ICCV. (2013)

\bibitem{UTInteractionData}
Ryoo, M.S., Aggarwal, J.K.:
\newblock {UT}-{I}nteraction {D}ataset, {ICPR} contest on {S}emantic
  {D}escription of {H}uman {A}ctivities ({SDHA}).
\newblock http://cvrc.ece.utexas.edu/SDHA2010/Human\_Interaction.html (2010)

\bibitem{soomro2012ucf101}
Soomro, K., Zamir, A.R., Shah, M.:
\newblock Ucf101: A dataset of 101 human actions classes from videos in the
  wild.
\newblock arXiv:1212.0402 (2012)

\bibitem{kuehne2011hmdb}
Kuehne, H., Jhuang, H., Garrote, E., Poggio, T., Serre, T.:
\newblock Hmdb: a large video database for human motion recognition.
\newblock In: ICCV. (2011)

\bibitem{kingma2015adam}
Kingma, D.P., Ba, J.:
\newblock Adam: A method for stochastic optimization.
\newblock ICLR (2015)

\bibitem{ryoo2010overview}
Ryoo, M., Chen, C.C., Aggarwal, J., Roy-Chowdhury, A.:
\newblock An overview of contest on semantic description of human activities
  (sdha) 2010.
\newblock In: ICPR.
\newblock (2010)

\bibitem{laviers2009improving}
Laviers, K., Sukthankar, G., Aha, D.W., Molineaux, M., Darken, C.,  et~al.:
\newblock Improving offensive performance through opponent modeling.
\newblock In: AIIDE. (2009)

\end{thebibliography}
